\pdfoutput=1

\documentclass[11pt]{article}

\usepackage{acl}

\usepackage{times}
\usepackage{latexsym}

\usepackage[T1]{fontenc}

\usepackage[utf8]{inputenc}

\usepackage{microtype}

\usepackage{algorithmic}
\usepackage[]{algorithm2e}
\usepackage{placeins}
\usepackage{xcolor}
\usepackage{comment}
\usepackage{amsmath}
\usepackage{multirow}
\usepackage{tabularx,booktabs}

\newcommand{\onetoone}{$\textsc{One2One}$}
\newcommand{\onetoseq}{$\textsc{One2Seq}$}

\title{Applying a Generic Sequence-to-Sequence Model for Simple and Effective Keyphrase Generation}

\author{
Md Faisal Mahbub Chowdhury,
    Gaetano Rossiello,
    Michael Glass, \\
    \bf Nandana Mihindukulasooriya \and
    Alfio Gliozzo\\
IBM Research AI, Yorktown Heights, NY, USA\\
\{mchowdh, mrglass, gliozzo\}@us.ibm.com, \\ \{gaetano.rossiello, nandana.m\}@ibm.com
}

\begin{document}
\maketitle

\begin{abstract}
In recent years, a number of keyphrase generation (KPG) approaches were proposed consisting of complex model architectures, dedicated training paradigms and decoding strategies. In this work, we opt for simplicity and show how a commonly used seq2seq language model, BART, can be easily adapted to generate keyphrases from the text in a single batch computation using a simple training procedure. Empirical results on five benchmarks show that our approach is as good as the existing state-of-the-art KPG systems, but using a much simpler and easy to deploy framework.
\end{abstract}

\section{Introduction}

Keyphrases are a collection of salient terms that summarize a document. Keyphrase generation (KPG) is unique from other NLG tasks since the target model prediction is to generate multiple sequences (i.e. a set of multi-word phrases) rather than a single sequence (for other tasks such as summarization and translation) \cite{meng-naacl-2021}. 

As pointed out by \citet{meng-naacl-2021}, the ever-growing collection of KPG studies, albeit novel, are overwhelming from their model design to hyper-parameter selection. It is worth to mention that, as of now, \emph{none of the existing KPG approaches} achieves outright state-of-the-art results 
on all of the five widely used keyphrase benchmarks.

\citet{cano-naacl-2019} experimented with popular text summarization neural architectures on several datasets. They found, despite using large quantities of data and many days of computation, the advanced deep learning based summarization models could not produce better keyphrases than the specifically designed keyphrase generation and extraction models. In this work, we revisit the above finding regarding the effectiveness of summarization models for keyphrases.

\begin{table}[t]
    \centering
    \small
    \def\arraystretch{1.2}
    \begin{tabular}{p{0.95\linewidth}}
        \textbf{Document} (only abstract shown below)\\\hline
        In order to monitor a region for traffic traversal, sensors can be deployed to perform \textbf{collaborative target detection}. Such a \textbf{sensor network} achieves a certain level of detection performance with an associated cost of deployment. This paper addresses this problem by proposing \textbf{path exposure} as a measure of the goodness of a deployment and presents an approach for \textbf{sequential deployment} in steps. It illustrates that the cost of deployment can be minimized to achieve the desired detection performance by appropriately choosing the number of sensors deployed in each step. \\\hline\hline
        \textbf{Generated present keyphrases}: target detection, path exposure, sensor network, collaborative target detection, sequential deployment \\\hline
        \textbf{Generated absent keyphrases}: traffic monitoring, sensor networks, sensor deployment, deployment strategy  \\\hline
    \end{tabular}
    \caption{An example of keyphrases generated using our fine-tuned model of BART-large~\cite{lewis-etal-2020-bart}. The document is selected from the SemEval dataset.}
    \label{tab:demo}
\end{table}

Recently, some proposed works  \cite{DBLP:conf/nips/LewisPPPKGKLYR020,petroni-etal-2021-kilt} show that a multitude of downstream NLP tasks (e.g. question answering, fact checking, dialog) can be effectively addressed using a unified generative framework without designing ad-hoc architectures for individual task.
Moreover, standard seq2seq (aka encoder-decoder) models~\cite{lewis-etal-2020-bart,2020t5} have been successfully applied to extract structured information from text, such as relation linking~\cite{DBLP:conf/semweb/RossielloMABGNK21}, text-to-table~\cite{DBLP:journals/corr/abs-2109-02707}, text-to-RDF~\cite{agarwal-etal-2020-machine, guo-etal-2020-2}, intent classification~\cite{ahmad-etal-2021-intent}, and slot filling~\cite{DBLP:conf/emnlp/GlassRCG21}. So far, there is no empirical evidence showing effectiveness of such a simple generative paradigm for KPG.

Inspired by the aforementioned literature, this study examines whether a standard text summarization neural architecture can be used for KPG. We demonstrate that, even with default hyper-parameters and limited fine tuning, our system can yield results on par (in some cases even better) with the state-of-the-art systems that have specifically designed architectures for KPG. We hope this study will have a substantial impact on implementing real-world KPG systems, therefore facilitating simplification of the development and deployment process.

\section{Related Work}
\label{sec:related_work}
KPG maps a sequence (the text) to a set of sequences (the keyphrases). This task is typically transformed to a seq2seq setting either through \onetoone{} or \onetoseq{}. In \onetoone{}, during training each keyphrase is paired with the input text as the target output. Then at test time the different keyphrases are taken from the different beams of the generation process. In contrast, \onetoseq{} uses the list of keyphrases as the target text during training. At test time the keyphrases are taken from the top-scoring beam or merged from the top-k beams.

\citet{meng-acl-2017} was the first to propose a KPG approach, called CopyRNN. They analyzed a number of benchmarks and found a considerable number (ranges between 44--58\%) of keyphrases do not appear in the corresponding documents. Such keyphrases are labelled as \emph{absent keyphrases} as opposed to the \emph{present keyphrases} that appear in the documents. CopyRNN is based on \onetoone{} training paradigm. Several subsequent studies \cite{chen-EMNLP-2018, zhao-zhang-acl-2019} are extensions of CopyRNN framework. Some \cite{ye-wang-emnlp-2018,chen-aaai-2019,kim-emnlp-2021} exploited document structure information to improve KPG performance.

\citet{yuan-acl-2020} implemented an approach based on \onetoseq{}  paradigm to generate diverse keyphrases and control number of outputs by manipulating decoder hidden states.

Unlike the aforementioned studies, \citet{chen-acl-2020} and \citet{chan-acl-2019} proposed reinforcement learning (RL) based KPG approaches. \citet{chen-NAACL-2019} proposed multi-task learning framework to jointly learn extractive and generative model for keyphrases and exploited information retrieval. They did not report separated results for present and absent keyphrases.

\citet{ye-emnlp-2021} proposed a graph neural network based approach that capture explicit knowledge from a predefined index (e.g., the training set) that is similar to the input document. This idea is intuitively similar to what was proposed by \citet{chen-NAACL-2019}.  Since we do not use explicit knowledge from other documents, these results are not comparable to ours.

\section{Proposed Approach}
\label{sec:proposed_approach}

We exploit BART \cite{lewis-acl-2020}, a denoising self-supervised autoencoder, for pretraining seq2seq models. It is a standard seq2seq framework that has been used to produce state-of-the-art results on a variety of NLG tasks including dialogue, question answering, abstractive summarization and translation. It has a bidirectional encoder (like BERT) and a left-to-right decoder (like GPT). BART uses a noise-added source text (e.g. by corrupting some tokens in source text) as input. It later uses language model for reconstructing the original text by predicting the true replacement of corrupted tokens.

There are two pretrained generic BART models, dubbed as bart-base and bart-large. They are made publicly available by ~\citet{wolf-etal-2020-transformers}.

\subsection{Proposed training procedure}
Let us define $D={\{w_{1}, \dots, w_{n}\}}$ as a document containing $n$ words, and $K={\{k_{1}, \dots, k_{m}\}}$ as a set consisting of $m$ keyphrases\footnote{Keyphrases can be represented as one or more words.}. We frame the KPG task as a sequence-to-sequence problem by fine-tuning BART using its conditional generation method typically used, e.g., for text summarization.

For this purpose, we need to define the representations for both source ($S$) and target ($T$) sequences.
$S$ is represented as a sequence of WordPieces obtained by applying the BART tokenizer on the text in $D$.
In order to create the sequence $T$, we organize all the keyphrases in $K$ in the following format: \texttt{[k\textsubscript{1}, k\textsubscript{2}, ..., k\textsubscript{m}]}. We organize the keyphrases in the same order as given in the ground truth. $T$ is obtained by applying the BART tokenizer on this sequence. This representation allows us to fine-tune the model by generating all the keyphrases jointly in a single decoding computation. Moreover, the characters (\texttt{[],}) are used as special tokens allowing us to easily parse the target sequence in order to extract the single keyphrases later during the inference phase.
Then, BART is fine-tuned by maximizing the probability $P(T|S) = \prod_{i=1}^{m} P(t_{i}| t_{<m},S)$ across all the instances in the given training set. We deliberately avoided adding new special tokens, e.g. to separate the keyphrases in the target sequence, in order to keep the model as simple as possible, thus promoting the usage of off-the-shelf training procedures provided in the Huggingface library.

It is worth noting that BART cannot handle documents having more than 1024 tokens. To address this issue, we split the input document into multiple paragraphs.
We split\footnote{Using NLTK - \url{https://www.nltk.org/}} the document into sentences and construct paragraphs by merging consecutive sentences until the maximum length accepted by the encoder is reached. Then, we consider the keyphrases provided as ground truth for the original document as the ground truth for each of these paragraphs. With this strategy, we treat each paragraph as an independent training instance.

\subsection{Proposed inference procedure}
Let's assume given an input test document, the system is asked to generate up to $N$ keyphrases. First, like in the training procedure, the document is split into paragraphs. Second, the keyphrase model is used to generate ranked list of $N$ keyphrases per paragraph.
Let $\mathcal{K}(p)$ be the ordered list of keyphrases generated for the paragraph $p$. The rank of each keyphrase $\kappa$ in $\mathcal{K}(p)$ is given by $rank(\kappa, \mathcal{K}(p))$.
Then the $score$ for a keyphrase is the sum of its inverse ranks over all the paragraphs in the document. We add $\epsilon = 1 / (N+1)$ to break score ties by the number of times a keyphrase is generated.
\begin{align*}
    score(\kappa) = \sum_{p_i \in D | \kappa \in \mathcal{K}(p_i)} \frac{1}{rank(\kappa, \mathcal{K}(p_i))}  + \epsilon
\end{align*}
Finally the top scoring $N$ keyphrases are selected as the final keyphrases for the document.

\section{Experiments}
\label{sec:experiments}

\subsection{Datasets}

We use the following 5 keyphrase benchmarks consisting of scientific publications for evaluation that have been widely used in existing literature - \texttt{Inspec} \cite{hulth-emnlp-2003}, \texttt{NUS} \cite{nguyen-2007-nus}, \texttt{SemEval} \cite{kim-semeval-2010}, \texttt{Kp20K} \cite{meng-acl-2017} and \texttt{Krapivin} \cite{Krapivin09largedataset}. We downloaded them from \url{https://github.com/memray/OpenNMT-kpg-release} shared by \citet{meng-naacl-2021}.  Due to space limitation, we refer the readers to \citet{meng-naacl-2021} or \citet{yuan-acl-2020} for various details and statistics about these datasets. All of these datasets have a test and validation data split. \texttt{Kp20K} also has a train split containing 500K+ abstracts. \texttt{SemEval}, \texttt{Krapivin} and  \texttt{NUS} also contain full texts in addition to abstracts.

\begin{table*}[t]
\small
\centering
\def\arraystretch{1.2}
\begin{tabularx}
{\textwidth}{r|XX|XX|XX|XX|XX}

  \multirow{2}{*}{\bf  Model} &
  \multicolumn{2}{c|}{\textbf{Inspec}} & \multicolumn{2}{c|}{\textbf{NUS}} &
  \multicolumn{2}{c|}{\textbf{SemEval}} &
  \multicolumn{2}{c|}{\textbf{Kp20K}} & \multicolumn{2}{c}{\textbf{Krapivin}}
  \\
  & F@5 & F@10 & F@5 & F@10 &  F@5 &   F@10 &  F@5 &  F@10 &  F@5 &   F@10
  \\ \hline

 ExHiRD-h~\cite{chen-acl-2020} & 25.3 & - & - & - & 28.4 & - & 31.1 & - & 28.6
 \\
 Transformer~\cite{chen-acl-2020} & 21.0 & - & - & - & 25.7 & - & 28.2 & - & 25.2
 \\

 CopyRNN~\cite{meng-acl-2017}  & 29.2 & 33.6  & 34.2 & 31.7 & 29.1 & 29.6 & 32.8 & 25.5 & 30.2 & 25.2
 \\

 CorrRNN~\cite{chen-EMNLP-2018} & - & - & 35.8 & 33.0 & 32.0  & 32.0  & - & - & 31.8 & 27.8
 \\

 ParaNetT+CoAtt~\cite{zhao-zhang-acl-2019} & 29.6 & 35.7 & 36.0 & 35.0 & 31.1 & 31.2 & \bf  36.0 & 28.9 & \bf  32.9 & 28.2
 \\

 catSeqTG-2RF~\cite{chan-acl-2019} & 25.3 & - & \bf  37.5 & - & 28.7 & - & 32.1 & - & 30.0 & -
 \\

 CatSeq~\cite{yuan-acl-2020} & 29.0 & 30.0 & 35.9 & 34.9 & 30.2 & 30.6 & 31.4 & 27.3 & 30.7 & 27.4
 \\

 CatSeqD~\cite{yuan-acl-2020} & 27.6 & 33.3 & 37.4 & 36.6 & \bf  32.7 & \bf  35.2 & 34.8 & 29.8 & 32.5 & \bf  28.5
 \\

 CatSeq+2RL(FB)~\cite{luo-emnlp-2021} & 26.7 & - & - & - & - & - & 33.0 & - & 30.5 & -
 \\
 GSEnc~\cite{kim-emnlp-2021} & - & - & - & - & - & - & 32.9 & - & -
 \\

\hline

 bart-base-kp (our) & 33.1 & 35.6 & 33.4 & 35.3 & 28.5 & 31.1 & 32.8 & 30.9 & 27.8 & 25.8
 \\

 bart-large-kp (our) & \bf  35.2 & \bf  38.7 & 34.6 & \bf  38.0 & 29.3 & 32.3 & 33.1 & \bf  31.1 & 26.3 & 25.0
 \\

\end{tabularx}
\caption{\label{tab:results_present_kp}\centering Evaluation for present keyphrases.}
\end{table*}

\subsection{Keyphrase model training}
\label{subsec:train_model}
We follow a simple recipe for training to avoid over-optimizing the models simply for the sake of obtaining the best possible results. For this reason, we fine-tune the conditional generation task in BART using a set of default hyper-parameters.\footnote{\emph{(i)} Epochs: 3, \emph{(ii)} learning rate: 5e-05, \emph{(iii)} train batch size: 128, \emph{(iv)} Adam beta: 0.9, \emph{(v)} Adam epsilon: 1e-08.}

With a small number of epochs (3), we use the train and validation splits in \texttt{Kp20K} to train a single keyphrase model using \texttt{BART-base} and another single model using \texttt{BART-large}. Henceforth, we will refer to them as \emph{bart-base-kp} and \emph{bart-large-kp}. Unlike some previous approaches (e.g. \citet{yuan-acl-2020}), we do not do any pre-processing (e.g. lowercasing, digits normalization, etc)  of either the document texts or ground-truth.

Also, we do not fine tune the \emph{bart-base-kp} and \emph{bart-large-kp} models for the other 4 non-Kp20K benchmarks using their corresponding validation split.

We used a machine with the following configuration for the training -- {\small \texttt{Processor:} Intel(R) Xeon(R) CPU E5-2690 v4 @ 2.60GHz, \texttt{CPU cores:} 16,  \texttt{RAM:} 118G, \texttt{GPU:} Tesla V100 PCIe 16G, \texttt{GPU count:} 2}. It took 21.82 hours to train the \emph{bart-base-kp} model and 51.68 hours for \emph{bart-large-kp}.

\begin{table*}
\centering
\small
\def\arraystretch{1.2}
\begin{tabular}{r|c|c|c|c|c}
  \multirow{2}{*}{\bf  Model} &
  \multicolumn{1}{c|}{\textbf{ Inspec}} & \multicolumn{1}{c|}{\textbf{ NUS}} &
  \multicolumn{1}{c|}{\textbf{ SemEval}} &
  \multicolumn{1}{c|}{\textbf{ Kp20K}} & \multicolumn{1}{c}{\textbf{ Krapivin}}\\
  &  R@10 &  R@10 &  R@10 & R@10 &  R@10 \\ \hline

 CopyRNN~\cite{meng-acl-2017} &  5.1 & 7.8 & \bf 4.9 & 11.5 & 11.6 \\

 CorrRNN~\cite{chen-EMNLP-2018}  &  - & 5.9 & 4.1 & - & - \\

 CatSeq~\cite{yuan-acl-2020} &  2.8 & 3.7 & 2.5 & 6.0 & 7.0 \\

 CatSeqD~\cite{yuan-acl-2020} & 5.2 & \bf 8.4 & 4.6 & \bf 11.7 & 12.0 \\
\hline
 bart-base-kp (our) & 4.8 & 5.6 & 3.0 & 6.1 & 11.2 \\

 bart-large-kp (our) & \bf 5.4 & 4.4 & 2.8 & 6.1 & \bf 13.2 \\

\end{tabular}
\caption{\label{tab:results_absent_kp}\centering Evaluation for absent keyphrases.}
\end{table*}

\subsection{Keyphrase prediction}

As noted by \citet{meng-naacl-2021}, most of the the existing approaches rely on beam search (with large beam size, e.g. 50, 150 and 200 in \cite{yuan-acl-2020, chen-aaai-2019, meng-acl-2017} respectively) to over-generate lots of keyphrases and then select the top N. This is not suitable in real-world situations that demand faster response with less memory footprint. In our experiments, we report results based on a default beam size \emph{20}.

We generate 10 keyphrases per source text, and hence report macro-average F-score@5 and F-score@10 for present keyphrases and macro-average Recall@10 for absent keyphrases over all the test documents in each benchmark.

We use the evaluation script of \cite{meng-naacl-2021}. It replaces punctuation marks with whitespace and tokenize texts/phrases using Python string method split() to reduce phrase matching errors. We used the concatenation of title and abstract as the source text and ignored full texts for the 3 benchmarks where they were available.

\section{Results and Analysis}
\label{sec:analysis}

\paragraph{Results:} As shown in Table \ref{tab:results_present_kp} and \ref{tab:results_absent_kp}, our model (\texttt{bart-large-kp}) achieves state-of-the-art \emph{F@10} in 3/5 and \emph{R@10} in 2/5 benchmarks for present and absent keyphrases respectively. Interestingly, our \texttt{bart-base-kp} model also produces very competitive results.

\paragraph{Qualitative Analysis:} Given the comparatively low performance of our system for absent keyphrases in \texttt{SemEval},
we examined 25\% randomly selected SemEval test documents and manually analyzed the absent keyphrases generated by the \texttt{bart-large-kp} model. 

On average, each of these selected documents contains 8.36 ground-truth absent keyphrases. But \texttt{bart-large-kp} generated only 2.05 absent keyphrases. So, recall was the main issue here. It should be noted that 36\% of the ground-truth in \texttt{Kp20K} (our training data) are absent keyphrases, but that number is 55\% in \texttt{SemEval} \cite{meng-naacl-2021}. Also, \texttt{SemEval} has 3 times more ground-truth per document than \texttt{Kp20K}.

We noticed 64\% of the total generated absent keyphrases in the above selected documents were found correct by the manual human analysis. But, 58\% of them were not in ground-truth, and could be considered as false negatives.

Regarding the generated absent keyphrases that we deemed incorrect (aka true negatives), there was a common pattern of errors where a number of absent keyphrases have spurious repetition of some tokens. For example, \textit{malmalware}, \textit{datadata transfer}, \textit{pairpairwise key establishment}, etc.

assessment but were not part of the gold standard). It highlights the challenge of formulating gold standards for absent/generative keyphrases as the number of possibilities are quite large.

\section{Conclusions}

\citet{boudin-acl-2020-keyphrase} showed that KPG can significantly boost effectiveness of information retrieval. But through extrinsic evaluation, they found existing KPG models have limited generalization ability.
In this work, we proposed an effective KPG system implemented using the same standard procedures used to fine-tune seq2seq language models on sequence generation tasks.
This strategy helps generalization since pre-trained transformer models, such as BART, promote transferability across domains.
In our experiments, we showed that the same model fine-tuned on a single dataset can be competitive also on other benchmarks compared with the existing state-of-the-art KPG models.
Moreover, its design simplicity is beneficial for deployment in real-world scenarios.

\bibliography{anthology, citations}
\bibliographystyle{acl_natbib}

\end{document}